\PassOptionsToPackage{table}{xcolor}

\documentclass[10pt,twocolumn,letterpaper]{article}

 \usepackage[pagenumbers]{cvpr} 

\usepackage[dvipsnames]{xcolor}

\definecolor{cvprblue}{rgb}{0.21,0.49,0.74}
\usepackage[pagebackref,breaklinks,colorlinks,citecolor=cvprblue,linkcolor=OrangeRed]{hyperref}

\usepackage[utf8]{inputenc} 
\usepackage[T1]{fontenc}    
\usepackage{hyperref}       
\usepackage{url}            
\usepackage{booktabs}       
\usepackage{amsfonts}       
\usepackage{nicefrac}       
\usepackage{microtype}      
\usepackage{graphicx}

\usepackage{amsthm,amsmath,amssymb}
\usepackage{mathrsfs}

\usepackage{multirow}

\usepackage{bm}
\usepackage{bbding}

\usepackage{url}
\usepackage{booktabs}

\usepackage{pifont}
\usepackage{wrapfig}
\usepackage{amssymb}
\usepackage{multirow}
\usepackage{bm}
\usepackage{bbding}

\usepackage{url}
\usepackage{booktabs}

\usepackage[table]{xcolor}

\def\paperID{2136} 
\def\confName{CVPR}
\def\confYear{2024}

\title{Adapter is All You Need for Tuning Visual Tasks}

\author{Dongshuo Yin$^{1,}\footnotemark[1] $ , Leiyi Hu$^{1,}\footnotemark[1] $ , Bin Li$^{2} $ , Youqun Zhang$^{2} $\\
	$^1$University of Chinese Academy of Sciences, $^2$Alibaba Group\\
	{\tt\small dongshuoyin@gmail.com, hu\_leiyi@163.com, $\{$zhuyi.lb, youqun.zhangyq$\}$@alibaba-inc.com}
}

\begin{document}
\maketitle
\renewcommand{\thefootnote}{\fnsymbol{footnote}}
\footnotetext[1]{Equal contribution.}

\begin{abstract}
Pre-training \& fine-tuning can enhance the transferring efficiency and performance in visual tasks. Recent delta-tuning methods provide more options for visual classification tasks. Despite their success, existing visual delta-tuning art fails to exceed the upper limit of full fine-tuning on challenging tasks like instance segmentation and semantic segmentation. To find a competitive alternative to full fine-tuning, we propose the \textbf{M}ulti-c\textbf{o}g\textbf{n}itive Visual \textbf{A}dapter \textbf{(Mona)} tuning, a novel adapter-based tuning method. First, we introduce multiple vision-friendly filters into the adapter to enhance its ability to process visual signals, while previous methods mainly rely on language-friendly linear filters. Second, we add the scaled normalization layer in the adapter to regulate the distribution of input features for visual filters. To fully demonstrate the practicality and generality of Mona, we conduct experiments on multiple representative visual tasks, including instance segmentation on COCO, semantic segmentation on ADE20K, object detection on Pascal VOC, and image classification on several common datasets. Exciting results illustrate that Mona surpasses full fine-tuning on all these tasks and is the only delta-tuning method outperforming full fine-tuning on instance segmentation and semantic segmentation tasks. For example, Mona achieves a 1\% performance gain on the COCO dataset compared to full fine-tuning. Comprehensive results suggest that Mona-tuning is more suitable for retaining and utilizing the capabilities of pre-trained models than full fine-tuning. The code will be released at \hyperlink{https://github.com/Leiyi-Hu/mona}{https://github.com/Leiyi-Hu/mona}.

\end{abstract}

\section{Introduction}
Pre-training \& fine-tuning paradigm \cite{wang2022pre} can perform impressive transfer learning between homo-modal tasks, as has been demonstrated in computer vision (CV) \cite{liu2021swin, fang2023eva} and natural language processing (NLP) \cite{tufano2022using, min2023recent, tinn2023fine}. Pre-trained models are often trained by well-resourced and experienced teams with large amounts of clean data \cite{yin2023parameter, yin20231}. Exceptional pre-trained models can help hardware- and data-limited teams save plenty of training costs and train well-performing deep models on new tasks \cite{sarasaen2021fine, amisse2021fine, too2019comparative, kading2016fine}. In the era of large models, the efficiency of tuning pre-trained models is an important issue. Full fine-tuning \cite{liu2021swin, fang2023eva} has been widely used with great success in CV tasks, which tunes all parameters in the pre-trained backbone as well as additional task-specific heads/necks during the training process. Many impressive CV art (e.g., Swin \cite{liu2021swin}, EVA \cite{fang2023eva}, etc.) broaden the upper limit of visual tasks in this way. However, is full fine-tuning still the best way to fine-tune visual tasks now? Our answer is NO.

\begin{figure}
	\centering
	\includegraphics[scale=.79]{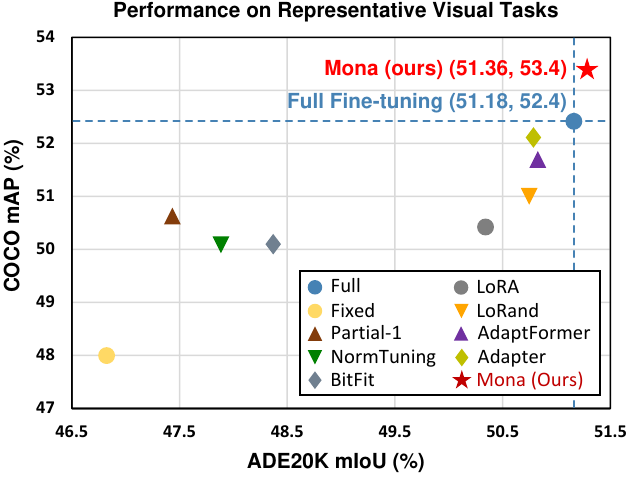}
	\caption{\textbf{Comparisons of our method with full fine-tuning and recent delta-tuning art on representative visual tasks.} Red dashed line is the performance of full fine-tuning on ADE20K and COCO. The proposed Mona outperforms full fine-tuning on representative visual tasks, which promotes the upper limit of previous delta-tuning art. The results demonstrate that the adapter-tuning paradigm can replace full fine-tuning and achieve better performance in most visual tasks. Full fine-tuning may no longer be the only preferred solution for transfer learning in the future.} \label{fig:first}
\end{figure}

Apart from full fine-tuning, Delta tuning \cite{ding2023parameter, hu2022sparse} has recently attracted attention in NLP and CV tasks. Delta tuning comes from NLP, which tunes only part of the backbone network or extra lightweight structures for efficient transfer learning \cite{ding2023parameter}. Delta tuning methods generally fix most backbone parameters and achieve comparable or even better performance than full fine-tuning on simple tasks (including classification tasks in NLP \cite{zhou2022making, rathnayake2022adapter} and CV \cite{jia2022visual, liupolyhistor, he2022parameter, chenadaptformer}). VPT \cite{jia2022visual} is the first to explore the potential of prompt-tuning on visual classification tasks. LoRand \cite{yin20231} pioneers adapter-tuning on dense predictions and reduces the gap between delta tuning and full fine-tuning on visual dense tasks. However, existing methods cannot outperform full fine-tuning on dense prediction tasks, including semantic segmentation and instance segmentation.

\begin{figure}
	\centering
	\includegraphics[scale=.8]{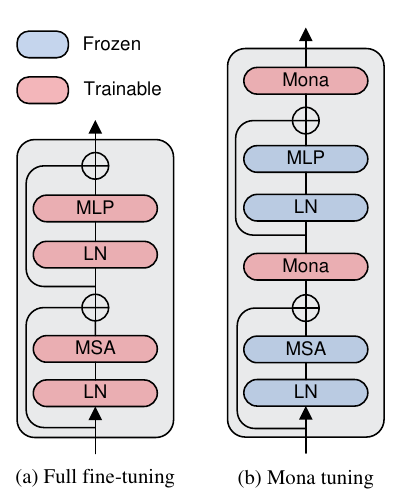}
	\caption{\textbf{Left:} All parameters in the classic full fine-tuning paradigm need to be updated. We employ the remarkable Swin Transformer series as backbones. \textbf{Right:} The proposed Mona-tuning. We add Mona after MSA and MLP in each SwinBlock. The proposed method fixes the parameters of pre-trained layers and updates the parameters of Mona.} \label{fig:adapter}
\end{figure}

To challenge the dominance of full fine-tuning in CV, we propose Mona-tuning, a novel tuning paradigm based on \textbf{M}ulti-c\textbf{o}g\textbf{n}itive visual \textbf{a}dapters \textbf{(Mona)}. We analyse recent art and summarise two problems in existing visual adapters. First, the designs of existing CV adapters \cite{liupolyhistor, he2022parameter, chenadaptformer} follow linear adapters in NLP \cite{houlsby2019parameter, hu2021lora}. Indeed, visual tasks process visual signals, which are significantly different from linguistic signals and have unique convolutional operations \cite{gu2018recent, li2021survey, albawi2017understanding}. Our experiments show that convolution-based filters can better transfer visual knowledge from pre-trained models to other tasks, so we propose a novel convolution-based adapter for visual tasks. Second, most existing adapters compress the upstream features into a single dimension and fit them to the new task via simple nonlinearity \cite{liupolyhistor, he2022parameter, chenadaptformer}. Previous works claim that models have different cognitions of features at different filter scales \cite{ozturk2018convolution, chansong2021impacts, agrawal2020using}. As a result, we employ multiple convolutional filters behind the adapter's reduction layer to enhance the cognitive abilities of the adapters. We demonstrate the generality and superiority of Mona-tuning on plenty of representative visual tasks, including image classification, object detection, semantic segmentation, and instance segmentation. We employ the SwinTransformer \cite{liu2021swin} series trained on ImageNet-22k \cite{deng2009imagenet} as pre-trained models. Extensive experiments indicate that the proposed method outperforms the traditional full fine-tuning paradigm both on simple image classification tasks and complex dense prediction tasks. For example, Mona-tuning outperforms full fine-tuning on the COCO dataset \cite{lin2014microsoft} by 1\% mAP. The results suggest that full fine-tuning may no longer be the optimal choice for visual tasks. Adapter-based tuning is a better-performing and more efficient paradigm for visual transfer learning. Moreover, Mona is the only method that surpasses full fine-tuning on semantic segmentation and instance segmentation. Figure \ref{fig:first} illustrates the superiority of the proposed method on the challenging instance segmentation and semantic segmentation tasks. Our contributions can be three-fold:

\begin{itemize}
\item We demonstrate that the adapter-based tuning can replace full fine-tuning on common visual tasks and achieve better performance with fewer new parameters.
\item We propose Mona-tuning, a novel training paradigm based on multi-cognitive visual adapters (Mona). Mona employs vision-friendly filters to optimise traditional linear adapters and improve visual transferring abilities through multiple cognitive perspectives.
\item Extensive experiments demonstrate that Mona-tuning outperforms full fine-tuning and other recent art on representative visual tasks, including image classification, object detection, semantic segmentation, and instance segmentation.
\end{itemize}

\section{Related Work}
\subsection{Delta-tuning}
The development of large models has produced dramatic shocks throughout artificial intelligence \cite{floridi2020gpt, touvron2023llama, kirillov2023segment}. The efficiency of transfer learning attracts researchers' interest \cite{chen2023sam, liupolyhistor, he2022parameter, chenadaptformer}. Delta tuning \cite{ding2023parameter, hu2022sparse, houlsby2019parameter, hu2021lora} (or parameter efficient fine-tuning PEFT) is dedicated to improving the efficiency of fine-tuning. Delta-tuning methods can be divided into three groups \cite{ding2023parameter}. The first group fixes most of the parameters in the pre-trained backbone and fine-tune a small number of them, e.g., BitFit \cite{zaken2021bitfit} tunes bias, Norm Tuning \cite{giannou2023expressive} tunes norm layers, and Partial-1 \cite{yosinski2014transferable} only tunes the last block. The second group reparameterises some parameters in the pre-trained model, e.g. the LoRA \cite{hulora} optimises low-rank subspaces. The third group fixes the pre-trained backbone's original parameters and adds additional trainable structures, including prompt series \cite{jia2022visual, liu2022p, zhu2023prompt} and adapter series \cite{sung2022vl, chen2022adaptformer, he2023parameter}. Our experiments compare Mona with these three groups.

\begin{figure}
	\centering
	\includegraphics[scale=.75]{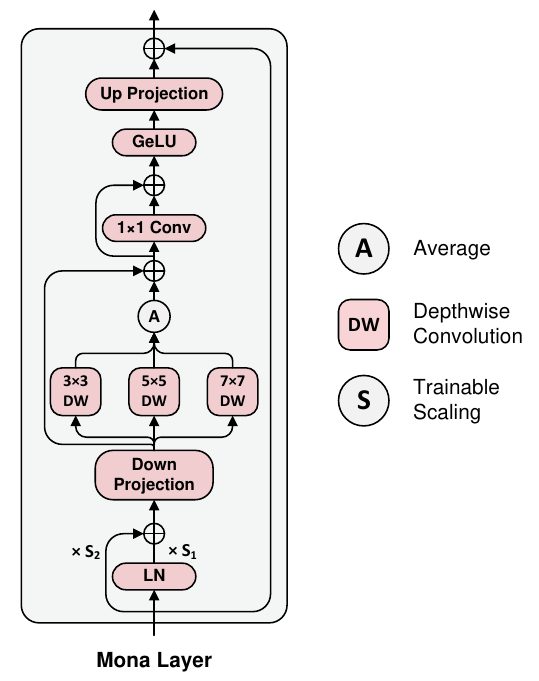}
	\caption{\textbf{Details of Mona.} Mona has a scaled LayerNorm before the down projection. A multi-cognitive convolutional filter group and an aggregation filter are behind the down projection. We add skip-connections at four places inside Mona to strengthen its adaptation capabilities. Mona enables the adapter-based fine-tuning paradigm to outperform full fine-tuning in typical visual tasks comprehensively.} \label{fig:mona}
\end{figure}

\subsection{Computer Vision Meets Delta-tuning}
Although derived from NLP, delta tuning is also explored in CV. VPT \cite{jia2022visual} is the first to introduce delta-tuning (prompt-tuning) to visual classification tasks. AdaptFormer \cite{chen2022adaptformer} designs a parallel adapter structure to improve delta-tuning performance on visual classification. KAdaptation \cite{he2023parameter} optimises the adapter through the Kronecker product. The above art is the pioneer in visual tasks, revealing the potential of delta-tuning on visual classification. LoRand \cite{yin20231} brings impressive performance on dense prediction tasks via multi-branch low-rank adapters but still cannot surpass full fine-tuning on all dense prediction tasks. Recent art indicates that delta-tuning cannot completely replace full fine-tuning on vision tasks. Therefore, we propose Mona-tuning, an alternative to full fine-tuning for more visual tasks, which outperforms full fine-tuning in both new parameter sizes and performance.

\begin{figure*}
	\centering
	\includegraphics[scale=.83]{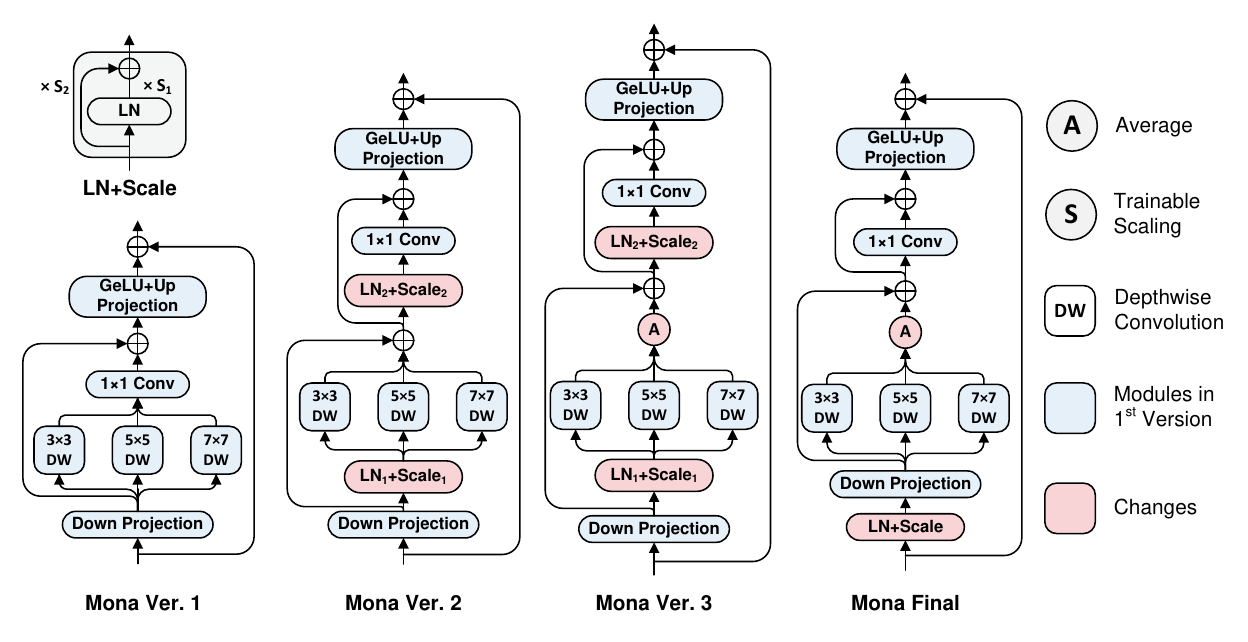}
	\caption{\textbf{Design Iterations.} Mona surpasses full fine-tuning on typical visual tasks after many iterations. The first version introduces multi-cognitive visual filters and outperforms previous adapter-based tuning art on visual tasks. We add LNs at different places and incorporate averaging operations to optimise Mona further. In fact, Mona finally succeeds in achieving our desired goal through more than 15 versions.} \label{fig:design}
\end{figure*}

\section{Methods}
In this section, we present the proposed method in four parts, including the adapter-tuning paradigm (Section \ref{adapter}), Mona (Section \ref{mona}), the design process (Section \ref{design}), and parameter analysis (Section \ref{param}).

\subsection{Adapter-tuning}
\label{adapter}
Previous work \cite{yin20231} discussed adapter fine-tuning, and we briefly introduce related concepts here. Figure \ref{fig:adapter} illustrates the difference between full fine-tuning (exemplified by SwinBlock) and the adapter-tuning paradigm (exemplified by Mona). Full fine-tuning updates all parameters in the pre-trained backbone, while adapter-tuning fixes the pre-trained parameters and updates the parameters in adapters. For dataset $D=\left\{(x_i,y_i)\right\}^N_{i=1}$, the optimization process of full fine-tuning and adapter-tuning can be expressed as Equation \ref{eq1} and Equation \ref{eq2}:

\begin{equation}
\label{eq1}
	\theta\gets\underset{\theta}{{\arg\min} \, loss(D,\theta)},
\end{equation}
\begin{equation}
\label{eq2}
\omega\gets\underset{\omega}{{\arg\min} \, loss(D,\theta_F, \omega)},
\end{equation}
where $loss$ is the training loss, $\theta$ represents parameters of the whole framework, and $\theta_F$ is the fixed parameters in adapter-tuning. $\omega$ represents updated parameters in adapter-tuning, including parameters in adapters and parameters outside the backbone.

\subsection{Mona}
\label{mona}
\noindent \textbf{Multi-Cognitive Visual Filters.}  Previous CV adapter art \cite{jia2022visual, liupolyhistor, he2022parameter, chenadaptformer} is based on linear structures, mainly including down projection, nonlinear activation, up projection, and skip connections. Vanilla adapter \cite{houlsby2019parameter} is for natural language signals and is not optimized for visual signals. Given the limitations of vanilla adapters, we focus on how to make the adapter better at transferring visual knowledge. For visual cognition, human eyes process visual signals from different scales and integrate them for better understanding \cite{koretz1988human, martinez2004role, szegedy2015going}. Adapters should also process upstream features from multiple cognitive perspectives for better performance on downstream tasks. We introduce multiple convolutional filters to Mona to increase the cognitive dimension. Depth-Wise Convolutions \cite{szegedy2015going} (DWConv) instead of standard convolutions are employed in Mona to minimize additional parameters. Specifically, the upstream features go through three DWConv filters after down projection. Convolution kernels are 3$\times$3, 5$\times$5 and 7$\times$7. We compute the average results from three filters and aggregate features with the 1$\times$1 convolution. Then, features are nonlinearized by GeLU. We add skip-connections at multiple places in Mona to minimize feature losses during convolution (see Figure \ref{fig:mona}). Finally, the feature dimension is recovered by up projection.

\noindent \textbf{Input Optimization.} For the adapter, the input features come from fixed layers. Fixed layers cannot adjust their parameter spaces according to new tasks' sample space, so the pre-training task affects inputs from the fixed layer. Vanilla adapter directly downscales inputs from fixed layers, which is not a good choice for visual tasks. As a result, we enable Mona to adjust the input distributions and the proportion of inputs from the fixed layer itself. Specifically, we add a norm layer and two learnable weights, $S_1$ and $S_2$, to the top end of Mona to adjust the input distribution. Previous work indicates that normalization \cite{xu2019understanding} helps to stabilize the forward input distribution and the backpropagated gradient. We find in practice that LayerNorm (LN) \cite{xu2019understanding} is better than BatchNorm \cite{ioffe2015batch}, so we employ LN in Mona. Figure \ref{fig:mona} illustrates our design.

\subsection{Design Process}
\label{design}
The design of Mona undergoes several iterations, as depicted in Figure \ref{fig:design}. The initial version incorporates multiple convolutional filters into the vanilla adapter to enhance its ability to process visual signals. This design outperforms previous adapter-based approaches and surpasses full fine-tuning in most tasks. We seek ways to optimise Mona to exceed full fine-tuning in all our tasks. Convolutional filters after down projection effectively enhance Mona's ability to alter visual knowledge in low-dimensional space. However, this version cannot optimise the input features originating from the fixed layers. We hope to send ``controllable'' and ``clean'' features to convolutional filters. As discussed in Section \ref{mona}, prior art demonstrates that LN can effectively adjust the distribution of features. In view of this, we try to enhance Mona by incorporating LN at various locations. Initially, we add Scaled LNs after down projection and before 1$\times$1 convolution ($2^{nd}$ version). However, this design does not yield results as promising as the initial version. We then average the summation of the DWConvs ($3^{rd}$ version), which slightly improves the second version's performance but still falls short of the first version. We consider that optimising the inputs of the subspace is not sufficient to improve the inputs of Mona. Therefore, we place the Scaled LN at the beginning of the entire adapter. In this version, Mona outperforms full fine-tuning across all tasks, and the final version also retains the positive averaging operation from the $3^{rd}$ version.

\subsection{Parameter Analysis}
\label{param}
The parameters of Mona come from LN, scaling factors, linear layers, DWconv and 1$\times$1 conv. Assuming that the input dimension of the adapter is $m$ and the dimension after down projection is $n$, the parameters of the LN and scaling factors are $2m+2$, the parameters of the two linear layers are $2mn+m+n$, the parameters of the DWConv layer are $(3^2+5^2+7^2)n=83n$, and the PWConv is $n^2$. The total parameter of each Mona module are $$(2n+3) m+n^2+84n+2.$$ For each block, all Mona parameters are: $2\times((2n+3)m+n^2+84n+2)$. We set the value of n to a constant (64) to reduce parameters in Mona.

\section{Experiments}
We implement sufficient experiments on multiple representative visual tasks to demonstrate the superiority of Mona-tuning. Sections \ref{dataset} $\sim$ \ref{basline} present the experimental settings, including datasets, pre-trained models and baselines. Section \ref{results} shows the experimental results and analysis. Section \ref{loss} analyses the convergence processes of different approches. We design ablation experiments in Section \ref{ablations} to illustrate some details and generalisability of the proposed method. Hyperparameters and detailed settings for model training are shown in the Supplementary Material.

\subsection{Datasets}
\label{dataset}
\noindent \textbf{Object Detection.} Pascal VOC 0712 \cite{everingham2015pascal} has 16k/5k training/validation images and is used for object detection tasks. We employ Swin-Large + RetinaNet for training. The evaluation metric for object detection task is the most commonly used AP$_{box}$.

\noindent \textbf{Semantic Segmentation.} ADE20K \cite{zhou2017scene} is the most widely used semantic segmentation dataset containing 20K training and 2K validation images. We employ Swin-Large + UperNet for experiments on semantic segmentation. The evaluation metric here is the most commonly used mIoU.


\noindent \textbf{Instance Segmentation.} MS COCO \cite{lin2014microsoft} is a representative instance segmentation dataset with 118k training images and 5k validation images. We employ Swin-Base + Cascade Mask RCNN for training. Evaluation metrics for instance segmentation task are AP$_{box}$ and AP$_{Mask}$.

\noindent \textbf{Image Classification.} Classification tasks have been well studied in previous art. To increase the broadness of our experiments, we also demonstrate Mona's generality on several widely used classification datasets. Specifically, we conduct our experiments on the Oxford 102 Flower Dataset \cite{nilsback2008automated}, the Oxford-IIIT Pet Dataset \cite{parkhi2012cats}, and the VOC 2007 Classification Challenge dataset \cite{pascal-voc-2007}. Oxford 102 Flower Dataset has 102 categories of flowers, with between 40 and 258 images per category. Oxford-IIIT Pet Dataset has 37 categories of pets, with about 200 images per category. VOC 2007 Classification Challenge Dataset contains about 10k images and 20 labeled categories. The top-1 and top-5 accuracies are used as evaluation metrics. We also report the average performance of each method.

\subsection{Pretrained models}
\label{pretrained}

The Swin Transformer series \cite{liu2021swin} is employed as the backbone for all experiments. The pre-trained models are trained on ImageNet-22k \cite{deng2009imagenet} and provided by OpenMMLab \cite{chen2019mmdetection}. The image resolution of the pretrained task is 224$\times$224. Experiments are performed on Nvidia Tesla V100s. Most tasks employ Swin-Large as the backbone. The backbone for COCO tasks is Swin-Base, considering the memory consumption of instance segmentation models.
\subsection{Baselines}
\label{basline}

We compare Mona with multiple recent methods. Baselines can be grouped into methods without or with extra structure:
\begin{itemize}
	\item Without extra structure:

%
%
%
%

\item[-] {\scshape Full}: Update all parameters in the framework.

\item[-] {\scshape Fixed}: Fix the backbone and update other parameters in the framework.

\item[-] {\scshape BitFit \cite{zaken2021bitfit}}: Update bias in backbone and parameters outside of backbone.

\item[-] {\scshape NormTuning \cite{giannou2023expressive}}: Update norm layers in backbone and parameters outside of backbone.

\item[-] {\scshape Partial-1 \cite{yosinski2014transferable}}: Update the last block in the backbone and parameters outside the backbone.

\end{itemize}

\begin{itemize}
\item With extra structure: 

%
%
%
%

\noindent (The pre-trained layers in these baselines are fixed, and the adapter intermediate dimensions are all 64, following the AdaptFormer \cite{chen2022adaptformer}):

\item[-] {\scshape Adapter \cite{houlsby2019parameter}}: Add standard adapter layers after the MSA/MLP layers of each SwinBlock.

\item[-] {\scshape LoRA \cite{hulora}}: Add parallel learnable matrices to multi-head attention weights.

\item[-] {\scshape AdaptFormer \cite{chen2022adaptformer}}: Add parallel adapter layers with scale weights to each MLP layer.

\item[-] {\scshape LoRand \cite{yin20231}}: Add LoRand layers after the MSA/MLP layers of each SwinBlock.

\end{itemize}

\begin{table*}[]
	\centering
	\scalebox{0.9}{
		\setlength{\tabcolsep}{3.5mm}{
			\begin{tabular}{@{}l|rrr|c|cccc@{}}
				\toprule
				\multirow{2}{*}{\textbf{\begin{tabular}[c]{@{}l@{}}\quad Swin-B\\ \quad  (89M)\end{tabular}}} & \multicolumn{1}{c}{\multirow{2}{*}{\textbf{\begin{tabular}[c]{@{}c@{}}Trained$\ast$ \\ Params\end{tabular}}}} & \multicolumn{1}{c}{\multirow{2}{*}{\textbf{\%}}} & 
				\multicolumn{1}{c|}{\multirow{2}{*}{\textbf{$\bm{\Delta_{Full}}$}}} &
				\multicolumn{1}{c|}{\multirow{2}{*}{\textbf{\begin{tabular}[c]{@{}c@{}}Extra\\ Structure\end{tabular}}}} & \multicolumn{4}{c}{\textbf{\begin{tabular}[c]{@{}c@{}}COCO\\ (Cascade Mask R-CNN)\end{tabular}}}  \\ \cline{6-9} 
				
				& \multicolumn{1}{c}{} & \multicolumn{1}{c}{} & \multicolumn{1}{c|}{} &  & \multicolumn{1}{c}{$\bm{\mathrm{AP_{Box}}}$} &\textbf{$\bm{\Delta_{Full}}$} & \multicolumn{1}{c}{$\bm{\mathrm{AP_{Mask}}}$} & \textbf{$\bm{\Delta_{Full}}$} \\ \midrule
				
				\rowcolor{gray!40}\multicolumn{9}{c}{\textit{\textbf{Baselines}}}\\ \midrule
				\quad {\scshape Full} & \multicolumn{1}{r}{ 89.14 M} & \multicolumn{1}{c}{100.00 \%} &\multicolumn{1}{c|}{-} & \multicolumn{1}{c|}{\ding{55}} & \multicolumn{1}{c}{ 52.40 \%} & \multicolumn{1}{c}{-} & \multicolumn{1}{c}{ 45.10 \%}  & -\\
				\quad {\scshape Fixed} & 0.00 M & 0.00 \% &- 100.00 \% & \ding{55} & \multicolumn{1}{c}{ 48.00 \%} & - 4.40 \%& 41.60 \% & - 3.50 \% \\ 
				\quad {\scshape BitFit} & 0.21 M & 0.23 \% & - 99.77 \%  & \ding{55} & \multicolumn{1}{c}{50.10 \%} & - 2.30 \% & 43.60 \% & - 1.50 \% \\
				\quad {\scshape NormTuning} & 0.06 M & 0.07 \% & - 99.93 \% & \ding{55} & \multicolumn{1}{c}{50.10 \%} & - 2.30 \% & 43.50 \%& - 1.60 \% \\ 
				\quad {\scshape Partial-1} & 12.95 M & 14.53 \% & - 85.47 \% & \ding{55} & \multicolumn{1}{c}{50.60 \%} & - 1.80 \% & 43.70 \%& - 1.40 \% \\ \midrule
				\quad {\scshape Adapter} & 3.19 M & 3.58 \% & - 96.42 \% & \ding{51} & \multicolumn{1}{c}{52.10 \%} & - 0.30 \% & 45.00 \%& - 0.10 \% \\ 
				\quad {\scshape LoRA} & 3.06 M & 3.43 \% & - 96.57 \% & \ding{51} & \multicolumn{1}{c}{50.40 \%} & - 2.00 \% & 43.90 \%& - 1.20 \% \\ 
				\quad {\scshape AdaptFormer} & 1.60 M & 1.79 \% & - 98.21 \% & \ding{51} & \multicolumn{1}{c}{51.70 \%} & - 0.70 \% & 44.60 \%& - 0.50 \% \\ 
				\quad {\scshape LoRand} & 1.20 M & 1.34 \% & - 98.66 \% & \ding{51} & \multicolumn{1}{c}{51.00 \%} & - 1.40 \% & 43.90 \%& - 1.20 \% \\ \midrule

				\rowcolor{gray!40}\multicolumn{9}{c}{\textit{\textbf{Our Method}}}\\ \midrule
				
				\quad {\scshape \textbf{Mona}} & 4.16 M & 4.67 \% & - 95.33 \% & \ding{51} & \textbf{53.4 \%} & \textbf{+ 1.00 \%}& \textbf{46.00} \% & \textbf{+ 0.90 \%} \\ \bottomrule
				
		\end{tabular}}
	}
	
	\caption{\textbf{Results of baselines and our methods on COCO benchmarks.} Swin-B is employed as the pre-trained model here. We present the numbers and percentages of trainable backbone parameters on the left and all the performences on the right. $\ast$ denotes the trainable parameters in backbones. The best AP in each column is bolded.}
	\label{tab:coco}
\end{table*}

\begin{table*}[]
	\centering
	\scalebox{0.9}{\setlength{\tabcolsep}{3.5mm}{
			\begin{tabular}{@{}l|rrr|c|cc|cc@{}}
				\toprule
				\multirow{2}{*}{\textbf{\begin{tabular}[c]{@{}l@{}}\quad Swin-L\\ \quad (198M)\end{tabular}}} & \multicolumn{1}{c}{\multirow{2}{*}{\textbf{\begin{tabular}[c]{@{}c@{}}Trained$\ast$ \\ Params \end{tabular}}}} & \multicolumn{1}{c}{\multirow{2}{*}{\textbf{\%}}} & 
				\multicolumn{1}{c|}{\multirow{2}{*}{\textbf{$\bm{\Delta_{Full}}$}}} &
				\multicolumn{1}{c|}{\multirow{2}{*}{\textbf{\begin{tabular}[c]{@{}c@{}}Extra\\ Structure\end{tabular}}}} & \multicolumn{2}{c|}{\textbf{\begin{tabular}[c]{@{}c@{}}Pascal VOC\\ (RetinaNet)\end{tabular}}} & \multicolumn{2}{c}{\textbf{\begin{tabular}[c]{@{}c@{}}ADE20K\\ (UperNet)\end{tabular}}} \\ \cline{6-9} 
				
				& \multicolumn{1}{c}{} & \multicolumn{1}{c}{} & \multicolumn{1}{c|}{} &  & \multicolumn{1}{c}{$\bm{\mathrm{AP_{Box}}}$} &\textbf{$\bm{\Delta_{Full}}$} & \multicolumn{1}{c}{$\bm{\mathrm{mIoU}}$} & \textbf{$\bm{\Delta_{Full}}$} \\ \midrule
				
				\rowcolor{gray!40}\multicolumn{9}{c}{\textit{\textbf{Baselines}}}\\ \midrule
				
				\quad {\scshape Full} & \multicolumn{1}{r}{ 198.58 M} & \multicolumn{1}{c}{100.00 \%} &\multicolumn{1}{c|}{-} & \multicolumn{1}{c|}{\ding{55}} & \multicolumn{1}{c}{83.70 \%} & \multicolumn{1}{c|}{-} & \multicolumn{1}{c}{ 51.18 \%}  & -\\
				\quad {\scshape Fixed} & 0.00 M & 0.00 \% &- 100.00 \% & \ding{55} & \multicolumn{1}{c}{ 83.80 \%} & + 0.10 \%& 46.84 \% & - 4.34 \% \\ 
				\quad {\scshape BitFit} & 0.30 M & 0.15 \% & - 99.85 \%  & \ding{55} & \multicolumn{1}{c}{85.40 \%} & + 1.70 \% & 48.37 \% & - 2.81 \% \\
				\quad {\scshape NormTuning} & 0.10 M & 0.05 \% & - 99.95 \% & \ding{55} & \multicolumn{1}{c}{85.50 \%} & + 1.80 \% & 47.89 \%& - 3.29 \% \\ 
				\quad {\scshape Partial-1} & 28.77 M & 14.53 \% & - 85.47 \% & \ding{55} & \multicolumn{1}{c}{85.50 \%} & + 1.80 \% & 47.44 \%& - 3.74 \% \\ \midrule
				\quad {\scshape Adapter} & 4.61 M & 2.33 \% & - 97.67 \% & \ding{51} & \multicolumn{1}{c}{86.70 \%} & + 3.00 \% & 50.78 \%& - 0.40 \% \\ 
				\quad {\scshape LoRA} & 4.57 M & 2.31 \% & - 97.69 \% & \ding{51} & \multicolumn{1}{c}{85.40 \%} & + 1.70 \% & 50.34 \%& - 0.84 \% \\ 
				\quad {\scshape AdaptFormer} & 2.34 M & 1.18 \% & - 98.82 \% & \ding{51} & \multicolumn{1}{c}{86.60 \%} & + 2.90 \% & 50.83 \%& - 0.35 \% \\ 
				\quad {\scshape LoRand} & 1.31 M & 0.66 \% & - 99.34 \% & \ding{51} & \multicolumn{1}{c}{86.80 \%} & + 3.10 \% & 50.76 \%& - 0.42 \% \\ \midrule

				\rowcolor{gray!40}\multicolumn{9}{c}{\textit{\textbf{Our Method}}}\\ \midrule
				
				\quad {\scshape \textbf{Mona}} & 5.08 M & 2.56 \% & - 97.44 \% & \ding{51} & \textbf{87.30 \%} & \textbf{+ 3.60 \%}& \textbf{51.36 \%} & \textbf{+ 0.18 \%} \\ \bottomrule
	\end{tabular}}}
	
	\caption{\textbf{Results of baselines and our methods on Pascal VOC and ADE20K benchmarks.} Swin-L is employed as the pre-trained model here. We present the numbers and percentages of trainable backbone parameters on the left and all the performences on the right. $\ast$ denotes the trainable parameters in backbones. The best AP/mIoU in each column is bolded.}
	\label{tab:ade}
\end{table*}

\begin{table*}[]
	\centering
	\scalebox{1.}{\setlength{\tabcolsep}{2mm}{
			\begin{tabular}{@{}lcccccccc@{}}
				\toprule
				\multirow{2}{*}{\quad \textbf{Method}} & \multicolumn{2}{c}{\textbf{Flowers102}}  & \multicolumn{2}{c}{\textbf{OxfordPets}}  & \multicolumn{2}{c}{\textbf{VOC2007} } & \multicolumn{2}{c}{\textbf{Average}} \\ \cmidrule(l){2-9} 
				& top-1 acc. & top-5 acc. & top-1 acc. & top-5 acc. & top-1 acc.     & top-5 acc. & top-1 acc.     & top-5 acc.    \\ \midrule
				
				\rowcolor{gray!30}\multicolumn{9}{c}{\textit{\textbf{Baselines}}}\\ \midrule
				
				\quad {\scshape Full}         &  99.5772  &  99.8536  &  94.6579  &  99.6257   &  84.1276  &  96.9507  & 92.7876 & 98.8100 \\
				\quad {\scshape Fixed}        &  99.3007  &  99.8374  &  94.2219  &  99.9182   &  85.0162  &  98.9499  & 92.8463 & 99.5685 \\
				\quad {\scshape BitFit}       &  99.5772  &  99.8211  &  95.3393  &  99.9182   &  85.6018  &  99.3336  & 93.5061 & 99.6910 \\
				\quad {\scshape NormTuning}   &  99.5284  &  99.8374  &  95.2303  &  99.8910   &  85.5210  &  99.2528  & 93.4266 & 99.6604 \\
				\quad {\scshape Partial-1}    &  99.6585  &  99.8374  &  95.3938  &  99.8637   &  84.9354  &  98.6066  & 93.3292 & 99.4359 \\
				\quad {\scshape Adapter}      &  99.5934  &  99.9024  &  95.3393  &  99.8092   &  \textbf{87.0355}  &  99.1317  & 93.9894 & 99.6144 \\
				\quad {\scshape LoRA}         &  99.5446  &  99.8536  &  95.1485  &  99.8910   &  85.7028  &  99.3134  & 93.4653 & 99.6860 \\
				\quad {\scshape AdaptFormer}  &  99.5609  &  99.8536  &  95.2576  &  99.8365   &  86.2884  &  99.2730  & 93.7023 & 99.6544 \\
				\quad {\scshape LoRand}       &  99.5531  &  99.8536  &  95.2315  &  99.8637   &  85.9526  &  99.1766  & 93.5791 & 99.6313 \\ \midrule
				\rowcolor{gray!30}\multicolumn{9}{c}{\textit{\textbf{Our Method}}}\\ \midrule
				\quad {\scshape \textbf{Mona}}&  \textbf{99.6764}  &  \textbf{99.8536}  &  \textbf{95.4765}  & \textbf{99.9182}  &  86.9709
				&  \textbf{99.5057} & \textbf{94.0413}&\textbf{99.7592 
				}\\ \bottomrule
				
	\end{tabular}}}
	\caption{\textbf{Results of baselines and our methods on three classification datasets.} Swin-L is employed as the pre-trained model here. We present top-1 accuracy (\%) and top-5 accuracy (\%) of each dataset. The best result in each column is bolded}
	\label{tab:cls}
\end{table*}

\subsection{Results}
\label{results}

Table \ref{tab:coco} shows the results of the proposed method and baselines on the COCO dataset. Instance segmentation is the most challenging task of all the experimental tasks, and COCO is also the largest of all the experimental datasets. Results on COCO can better demonstrate the potential of adapter-tuning in visual tasks compared to other tasks. From Table \ref{tab:coco}, we find that Mona, based on multi-cognitive visual filters, outperforms all baselines. Moreover, Mona is the only method that outperforms full fine-tuning, resulting in a significant gain of 1\%. COCO experiments effectively demonstrate the capability of the proposed method and show a better option than full fine-tuning in terms of storage and performance. Among delta-tuning methods, most baselines without extra structure can save more new parameters (except Partial-1), but their average performance is lower than that with extra structure. For baselines with additional structure, the adapter-based approach is superior to the reparameterization-based approach (LoRA). LoRA is one of the recent representative delta-tuning approaches widely used in NLP tasks, but its performance weaknesses render it unsuitable for computer vision tasks. Table \ref{tab:coco} indicates that the performance of delta-tuning is not directly proportional to parameter sizes. Partial-1 has the most updated parameters among all baselines, but its performance is significantly lower than that of adapter-based baselines. This result suggests that superior module design can effectively enhance the transferring efficiency of pre-trained models while reducing massive storage consumption.

We show results in Table \ref{tab:ade} for two tasks, namely object detection on Pascal VOC and semantic segmentation on ADE20K. The proposed Mona outperforms all baseline methods on these two representative vision tasks. Mona produces a performance gain of 3.6\% and 0.18\% on the two tasks compared to full fine-tuning. Table \ref{tab:ade} again indicates that full fine-tuning is not the best choice for visual transfer learning. For other baselines, conclusions on COCO are confirmed again on VOC and ADE20K. Interestingly, it is different from COCO and ADE20K that all baselines exceed full fine-tuning on VOC. The VOC dataset has relatively little data, which may lead to overfitting when full fine-tuning a 198M Swin-Large pretained model. Compared to full fine-tuning, other methods fix most pre-trained parameters, so the model performance is less likely to collapse severely during tuning. NLP scholars treat similar cases as low-resource cases \cite{dodge2020fine, peters2019tune}. The object detection results here can be considered as a low-resource case in CV. For ADE20K, the performance gaps between baselines without additional structure and adapter-based baselines are more significant than VOC and COCO. For parameter sizes, most methods in Tables \ref{tab:coco} and \ref{tab:ade} (except Partial-1) produce less than 5\% new backbone parameters, which is the characteristic and advantage of delta-tuning. Despite the slight increase in parameters, Mona still outperforms the previous art and breaks the full fine-tuning performance ceiling by a wide margin.

We have shown the individual and average results on three classification datasets in Table \ref{tab:cls}. Mona outperforms all the baselines on Flowers102, OxfordPets, and outperforms the average results of all baselines. Table \ref{tab:cls} indicates that Mona has a high transfer efficiency on relatively simple tasks. In addition, we find that the average results of all delta-tuning methods surpass full fine-tuning, which is similar to conclusions in previous art \cite{he2022parameter}. The pre-trained model we used is Swin-Large (198M parameters) trained on ImageNet-22K, whose powerful knowledge has enabled Flower102 and OxfordPets to achieve very high scores. Compared to classification tasks, more complex dense prediction tasks (object detection, semantic segmentation, instance segmentation) are more suitable for reflecting the differences between different fine-tuning paradigms.

In summary, the results of Tables \ref{tab:coco} to \ref{tab:cls} can be summarized in two aspects: 1) As to performance, the widely used full fine-tuning paradigm in art like Swin and EVA is no longer the optimal choice for visual tasks. The proposed Mona-tuning surpasses the performance ceiling of full fine-tuning in representative tasks such as instance segmentation, semantic segmentation, object detection, and image classification. Specifically, Mona achieves a 1\% AP gain over full fine-tuning in the challenging COCO instance segmentation task. 2) Mona, based on multi-cognitive visual filtering, surpasses recent remarkable baselines in most tasks. Mona comprehensively enhances the practicality and generality of delta-tuning in visual tasks. Mona-tuning not only significantly reduces storage costs, but also further elevates the performance ceiling of visual tasks.

\subsection{Loss Analysis}
\label{loss}

\begin{figure}
	\centering
	\includegraphics[scale=.39]{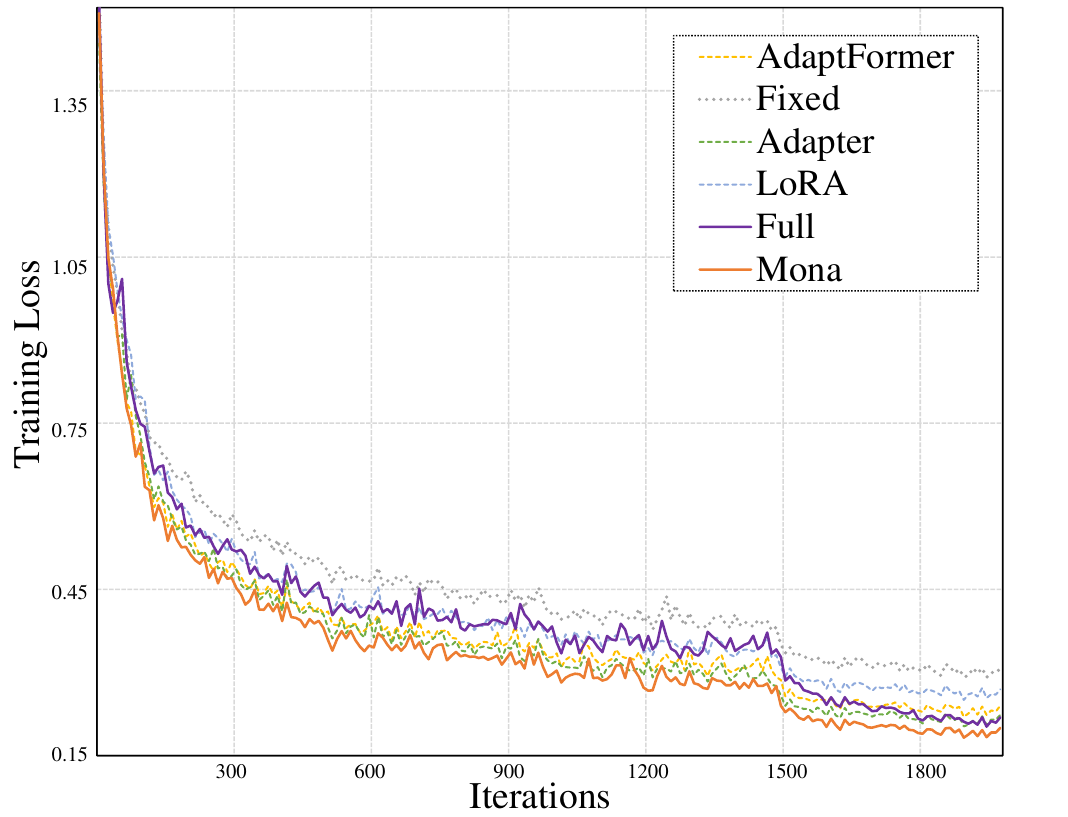}
	\caption{\textbf{Loss curves.} Among all the methods, the proposed method converges faster and significantly exceeds the full fine-tuning.} \label{fig:loss}
\end{figure}

We present the loss converging process for Mona and five representative  baselines on the object detection task (Pascal VOC) in Figure \ref{fig:loss}. The proposed method yields a significant advantage in the convergence process compared to full fine-tuning, which explains its better performance on VOC. Mona also converges faster than other delta-tuning methods, suggesting that multi-cognitive visual filters can better process visual features and accelerate the convergence of transfer learning. Convergence analysis again demonstrates that the proposed method is a highly competitive visual transfer learning method and full fine-tuning is no longer the optimal choice for visual tasks.

\subsection{Ablations}
\label{ablations}

In this section, we conduct several ablation experiments to discuss some detailed issues of the model, including the impact of intermediate dimensions on the model results and the relationship between model sizes and Mona-tuning. For the fairness and clarity of the experiment, all ablation experiments are conducted on Pascal VOC.

The workflow of the adapter is to compress the input from pre-trained layers into a low-dimensional feature space and transfer the knowledge in pre-trained layers to the new models by updating the adapter's parameters. Therefore, the intermediate dimension of the adapter is an important factor for model performance. We ablate the intermediate dimension of Mona and present the results in Table \ref{tab:ab-dim}. We only change the dimensions of Mona and fix other settings. Dimension candidates are 32, 64, and 128. The results of Table \ref{tab:ab-dim} show that the 64-dimension result surpasses that of smaller 32-dimension and larger 128-dimension, which is interesting. Chen et al. \cite{chen2022adaptformer} also study the intermediate dimension of AdaptFormer. They find that the 64-dimension AdaptFormer surpasses its 32- and 256-dimension versions in visual classification tasks, which is consistent with our conclusion. The results of Table \ref{tab:ab-dim} and Chen et al. indicate that the intermediate dimension of the adapter is not proportional to the performance, which means that a larger number of adapter parameters does not necessarily lead to better results. 

In addition to the above issue, we are also very concerned about Mona's performance on models of different sizes. We change the size of the backbone network under the same settings, and the model candidates are 29M Swin-T, 88M Swin-B, and 197M Swin-L. Table \ref{tab:ab-model} shows the results of full fine-tuning and Mona-tuning under three settings. We can draw the following three conclusions from Table \ref{tab:ab-model}. First, the more parameters the backbone network has, the smaller the proportion of Mona parameters for the same Mona setting. This result indicates that Mona-tuning can save more parameters when the backbone gets larger. Existing visual models are getting larger and larger. InternImage-H \cite{wang2023internimage} reaches 1.08B parameters, and SwinV2-G \cite{liu2022swin} reaches 3B. Parameter-efficient Mona-tuning can save billions of parameters and massive storage costs in the era of large models. Second, Mona surpasses full fine-tuning on three model settings, and its performance improves when model size grows. Table \ref{tab:ab-model} shows that Mona-tuning can improve training efficiency and performance in smaller models. We just discussed Mona's advantages for large models. However, more resource-limited research teams and project groups use small models. Mona-tuning also has the potential to help resource-limited visual researchers effectively leverage high-performance large models in their own applications. Third, the proposed method is more capable of stimulating the potential of large models compared to full fine-tuning. From Swin-T to Swin-L, full fine-tuning brings 3.6\% performance gain, while Mona brings 3.8\%. In other words, Mona can achieve better performance as the model gets larger and help further increase the upper bound for performance-sensitive tasks.

\begin{table}[]
	\centering
	\scalebox{.8}{\setlength{\tabcolsep}{3.9mm}{
	\begin{tabular}{@{}ccc@{}}
		\toprule
		\begin{tabular}[c]{@{}c@{}}\textbf{Intermediate} \\ \textbf{Dimensions}\end{tabular} & \multicolumn{1}{l}{\begin{tabular}[c]{@{}l@{}}\textbf{Trained}\\ \textbf{Params*}\end{tabular}} & $\bm{\mathrm{AP_{Box}}}$   \\ \midrule
		32   & 1.35 \%      & 86.8 \%\\
		64   & 2.56 \%      & 87.3 \%\\
		128  & 5.22 \%      & 87.1 \%\\ \bottomrule
	\end{tabular}}}
\caption{\textbf{Ablations of intermediate dimensions.} We fix other settings and evaluate the performance of Mona with 32, 64, and 128 intermediate dimensions on the VOC dataset. The configuration with 64 intermediate dimensions achieves the best performance and is consequently chosen as the general setting for Mona. $\ast$ denotes the trainable parameters in backbones.}
\label{tab:ab-dim}
\end{table}

\begin{table}[]
	\centering
	\scalebox{.8}{\setlength{\tabcolsep}{3.9mm}{
	\begin{tabular}{@{}lccc@{}}
		\toprule
		\textbf{Model} &\begin{tabular}[c]{@{}c@{}}{\scshape \textbf{Full}}\\ \textbf{(VOC)}\end{tabular} & \begin{tabular}[c]{@{}c@{}}{\scshape \textbf{Mona}}\\ \textbf{(VOC)}\end{tabular} & \multicolumn{1}{c}{\begin{tabular}[c]{@{}c@{}}\textbf{Param \%}\\ \textbf{(Mona)}\end{tabular}} \\ \midrule
		Swin-T & 80.1 \%& 83.5 \% & 4.87 \%                                                                        \\
		Swin-B & 81.6 \% & 86.5 \%& 4.06 \%                                                                        \\
		Swin-L & 83.7 \%& 87.3 \%& 2.56 \%                                                                        \\ \bottomrule
	\end{tabular}}}
	\caption{\textbf{Performance of mona on models with different sizes.} Mona outperforms full fine-tuning in all configurations, indicating that the model size does not constrain Mona's superiority. Mona can save more new parameters on larger models, which is significant in the era of large models.}
	\label{tab:ab-model}
\end{table}

\section{Conclusion}
This paper propose a novel visual fine-tuning method, the multi-cognitive visual adapter (Mona) tuning, which effectively enhances the efficiency and performance of visual fine-tuning. Comprehensive experiments demonstrate that the proposed Mona outperforms traditional full fine-tuning paradigms and other delta-tuning methods across representative tasks, including instance segmentation, semantic segmentation, object detection, and image classification. In the era of large models, full fine-tuning is no longer the optimal choice for visual tasks. We hope that Mona-tuning can effectively improve the transferring efficiency of large models and bring performance breakthroughs on more visual tasks.

{
    \small
    \bibliographystyle{ieeenat_fullname}
    \bibliography{ref}
}

\clearpage

\appendix{\noindent\textbf{Appendix}}
\\
\\
\noindent\textbf{A. Experimental Details}
\\
\\
\noindent\textbf{A.1 Code Implementation}

\noindent The implementations of semantic segmentation, instance segmentation (object detection) and classification are individually based on \textit{mmsegmentation}\footnote{https://github.com/open-mmlab/mmsegmentation}, \textit{mmdetection}\footnote{https://github.com/open-mmlab/mmdetection/} and \textit{mmpretrain}\footnote{https://github.com/open-mmlab/mmpretrain}. Parameters in Mona are initialized by Kaiming Uniform.
\\
\\
\noindent \textbf{A.2 Hyperparameters}

\begin{table}[h]
	\centering
	\scalebox{.92}{\setlength{\tabcolsep}{1mm}{
			\begin{tabular}{@{}lllllllll@{}}
				\toprule
				\textbf{Tasks}  & \textbf{Training Scheme} & \textbf{No. Epoch/Iter.} & \textbf{Optimizer} & \textbf{Learning Rate} & \textbf{Weight Decay} & \textbf{Batch Size} & \textbf{Mona lr Times} & \textbf{Backbone} \\ \midrule
				COCO            & By Epoch                 & 36                       & AdamW              & 0.0001                 & 0.05                  & 8                   & x1                     & Swin-B            \\
				VOC             & By Epoch                 & 12                       & AdamW              & 0.0001                 & 0.05                  & 2                   & x1                     & Swin-L            \\
				ADE20K          & By Iteration             & 160k                     & AdamW              & 0.00006                & 0.01                  & 16                  & x5                     & Swin-L            \\
				Classifications & By Epoch                 & 100                      & AdamW              & 0.00001562             & 0.05                  & 4                   & x2                     & Swin-L            \\ \bottomrule
			\end{tabular}
	}}
	\caption{Experiment settings of four tasks.}
\end{table}


\end{document}